\documentclass{article} 
\usepackage{nips14submit_e,times}
\usepackage{hyperref}
\usepackage{url}
\usepackage{amsmath}
\usepackage{algorithm}
\usepackage[noend]{algpseudocode}

\usepackage{graphicx}
\usepackage{caption}
\usepackage{subcaption}

\makeatletter
\def\BState{\State\hskip-\ALG@thistlm}
\makeatother

\usepackage{mathtools}
\DeclareMathOperator*{\argmin}{argmin}

\title{Mind the Gap: Subspace based Hierarchical Domain Adaptation}

\vspace{-0.6cm}

\author{
Anant Raj\thanks{Acknowledgement: The authors acknowledge support from ERC Project Cognimund and Research I Foundation, IITK} \\
IIT Kanpur\\
Kanpur, India 208016 \\
\texttt{anantraj@iitk.ac.in} \\
\And
Vinay P. Namboodiri \\
IIT Kanpur \\
Kanpur, India 208016 \\
\texttt{vinaypn@iitk.ac.in} \\
\AND
Tinne Tuytelaars \\
PSI-VISICS/ESAT \\
K.U. Leuven \\
Heverlee, Belgium 3001\\
\texttt{tinne.tuytelaars@esat.kuleuven.be} \\
}
\vspace{-0.6cm}

\nipsfinalcopy 

\begin{document}

\vspace{-0.5cm}
\maketitle

\vspace{-0.5cm}
\begin{abstract}
\vspace{-0.5cm}
Domain adaptation techniques aim at adapting a classifier learnt on a source domain to work on the target domain. Exploiting the subspaces spanned by features of the source and target domains respectively is one approach that has been investigated towards solving this problem. These techniques normally assume the existence of a single subspace for the entire source / target domain. In this work, we consider the hierarchical organization of the data and consider multiple subspaces for the source and target domain based on the hierarchy. We evaluate different subspace based domain adaptation techniques under this setting and observe that using different subspaces based on the hierarchy yields consistent improvement over a non-hierarchical baseline. 
\end{abstract}
\vspace{-0.6cm}
\section{Introduction}
\label{intro}
\vspace{-0.5cm}
While evaluating unseen test instances on a classifier trained over a set of labelled training instances, there is a standard assumption that test instances and training instances follow the same distribution. However, many real world scenarios violate this 
assumption. Think of a case where someone wants to classify the images taken with his low quality phone camera for which he doesn't have labels available. Can the person classify those images using the classifier which was trained on some publicly available dataset like ImageNet or Flickr ?
The obvious answer is no. Many studies have shown that if the test instances are not sampled from the same distribution as the training instances then the performance of the classifier significantly diminishes \cite{ES4,ES14,ES15}. This problem of domain shift is also extensively studied in the field of natural language processing and speech processing \cite{ES6,ES7}. To address this challenge, methods have been suggested to adapt a domain \textit{(Source Domain)} with respect to the other domain \textit{(Target Domain)} so that a classifier trained on \textit{Source Domain} data also contains the property of \textit{Target Domain} data. One can distinguish two settings in the domain adaptation literature: (1) \textit{the unsupervised setting} when the target domain is completely unlabeled and (2) \textit{the semi-supervised setting} when the target domain is partially labeled. In both 
settings, the source domain is fully labelled. In this work we focus on the unsupervised setting that is more challenging one. A promising line of work to solve this problem is by subspace based domain adaptation \cite{ES1,ES2,ES3}. However, none of the above approaches takes the semantic (dis)similarity of the category classes into account. Classes which are semantically similar have a very different distribution than classes which are semantically different. Based on this observation, we 
advocate that it's better to align the subspaces separately rather than considering the whole target data distribution at once. \\
To address this challenge we propose a new method of \textit{step-wise subspace alignment} for domain adaptation. Step-wise subspace alignment here indicates that we first align the subspaces for a set of larger clusters or group of semantically similar categories and then for the categories within the clusters. \\
We evaluate the effectiveness of proposed approach on a standard dataset having classes arranged according to their semantics in the hierarchy. However, the proposed approach can also be effective for the case when the hierarchy is not available. In such scenario similar categories can be clustered together in unsupervised way. 

\vspace{-0.4cm}
\section{Related Work}
\label{relwork}

\vspace{-0.4cm}
As mentioned in section \ref{intro} domain adaptation is widely studied in many fields including Natural Language Processing, Speech Processing and Computer Vision \cite{ES4,ES6,ES9}. A survey on recent advances in domain adaptation in natural language processing and computer vision can be found in \cite{ES5,ES8,ES13}. Subspace based approaches are most popular for solving the visual domain shift problem \cite{ES1,ES2,ES3}. The same principal lies behind these approaches. They first determine separate subspaces for source and target data and then project
the data onto these subspaces and/or a set of intermediate sampled subspaces with the aim of making the feature point domain invariant. In \cite{ES3}, a method is proposed to sample subspaces along the geodesic between source and target subspace on the Grassmann manifold. Once sampling is done then features are projected onto those sampled subspaces and a classifier is trained on the projected features. In \cite{ES2}, the geodesic flow kernel
is proposed to capture the incremental details in subspaces between source and target subspace along the geodesic. Instead of using intermediate
subspaces, \cite{ES1} proposes to learn a transformation to directly align the source subspace to the target subspace.  

Only few works have looked at the use of hierarchies in the context of domain adaptation. In \cite{ES10}, Nguyen {\it et al.} propose to adapt a hierarchy of features to exploit the richness of visual data.  The intent behind this work is similar to our work, in that semantic closeness and context information are exploited to boost
domain adaptation performance. Taking this idea forward a recent work on hierarchical adaptive structural SVM for domain adaptation has been proposed in \cite{ES11}. They organize multiple target domains into a hierarchical structure (tree) and adapt the source model to them jointly. Others have used statistical methods for hierarchical domain adaptation, e.g. in \cite{ES12} a hierarchical Bayesian prior is used to solve the domain shift problem in natural language and speech processing. However, the previous works have assumed a single common subspace between source and target, while our approach makes use of the hierarchical structure among the different classes to learn separate subspaces.

\vspace{-0.3cm}
\section{Background}
\label{background}
\vspace{-0.3cm}
 The proposed approach builds up on the previously proposed subspace based methods \cite{ES1,ES2, ES3}. One could learn the domain shift between source and target data on the original features itself. However this would be sub-optimal and involve significantly modifying the classifiers. Therefore, it is more common to learn it on a more robust representation of the data by first selecting $d$ dominating eigenvectors obtained using \textit{principal} \textit{component} \textit{analysis}. These $d$ eigenvectors work as the basis vectors for the \textit{source} and \textit{target} subspaces. The source and target features are then projected on the subspaces. Two recently proposed state-of-the-art subspace based domain adaptation methods \cite{ES1,ES2} used in this paper are discussed in \ref{subspace_alignment} and \ref{gfk}

\vspace{-0.3cm}
\subsection{Subspace Alignment}
\label{subspace_alignment}
\vspace{-0.3cm}
Subspace alignment based domain adaptation method consists of learning a transformation matrix $M$
that maps the source subspace to the target one \cite{ES1}. The mathematical formulation to this problem is given by 
\begin{eqnarray}
\vspace{-0.4cm}
F(M) = \|X_SM-X_T\|_F^2 \qquad  M^* = \argmin_M (F(M)).
\vspace{-0.4cm}
\label{equ1}
\end{eqnarray}
 $X_s$ and $X_t$ are matrices containing the $d$ most important eigenvectors for source and target respectively. $M$ is a transformation matrix from the source subspace $X_S$ to target subspace $X_T$ and $\|.\|_F$ is the \textit{Frobenius norm}. The solution of \textit{eq.} \ref{equ1} is $M^* = X_S'X_T$ and hence for the target aligned source coordinate system we get $X_a = X_SX_S'X_T$. 

\vspace{-0.3cm}
\subsection{Geodesic Flow Kernel}
\label{gfk}
\vspace{-0.3cm}
The geodesic flow kernel based domain adaptation method constructs an  infinite-dimensional feature space that carries the information of incremental change from source to target domain data \cite{ES2}. A key step in this method is to determine the geodesic curve between the two subspaces and to construct the geodesic flow kernel. If $X_S$ and $X_T$ are source and target subspaces having the same dimension then these two subspaces are separate points on a Grassmann manifold which is also a \textit{Riemannian Manifold}. Let $R_S$ be the orthogonal complement to $X_S$. From the property of \textit{Riemannian Manifold}, flow from $X_S$ towards $X_T$ can be calculated as: 
\begin{eqnarray}
\vspace{-0.2cm}
		\phi(t) = X_SU_1 \Gamma(t) - R_SU_2\Sigma(t) \mbox{ where } 
		 X_S^TX_T = U_1\Gamma V^T , R_S^TX_T = - U_2\Sigma V^T \nonumber.
\vspace{-0.3cm}
\end{eqnarray}
Based on the decomposition of the source subspace $X_S$ and its orthogonal complement $R_S$, we can obtain a geodesic flow kernel matrix $G$ that is given by\\
\vspace{-0.3cm}
$\left[ \begin{array}{cc} G \end{array} \right] = \left[ \begin{array}{cc} P_SU_1 & R_SU_2 \end{array} \right]  \begin{bmatrix} \Lambda_1 & \Lambda_2 \\ \Lambda_2 & \Lambda_3 \end{bmatrix}   \left[ \begin{array}{c} U_1^TP_S^T \\ U_2^TR_S^T \end{array} \right]$ where $\Lambda s$ are diagonal matrices that depend on the principal angles. 

Once we obtain the geodesic flow matrix $G$, we can relate labeled samples $x_i$ from the source subspace $X_i$ and unlabeled samples $x_j$ from the target subspace $X_j$ by using the distance metric $x_i^T [G] x_j$.

\vspace{-0.5cm}
\section{Our Approach}
\label{approach}
\vspace{-0.3cm}

 In this section we describe how the methods explained in section \ref{background} are adapted for hierarchical domain adaptation. Instead of using the same subspace throughout, we postulate that better results can be obtained by using different subspaces for different levels of the hierarchy. Indeed,
the more specific subspaces spanned by instances of categories of a certain branch of our tree (corresponding to similar categories), can be expected to better fit the data and therefore better model the domain shift. For the source domain, these subspaces can easily be obtained. For the target domain, however, no class labels are available as we are
working in the unsupervised setting. Therefore, the exact subspaces cannot be computed.
We circumvent this problem by first predicting the parent class label for each instance,
using the global subspaces and applying domain adaptation at the level of the root node.
We then use these predicted parent class labels to compute the next level of subspaces.
This results in a two step approach, as summarized in the algorithm below.
\vspace{-0.3cm}
\begin{algorithm}
\caption{Subspace Based Hierarchical Domain Adaptation}\label{alg:euclid}
\begin{algorithmic}[1]
\Procedure{Hierarchical Domain Adaptation}{Source Data S,Target Data T}
\State $X_{S_{root}} \gets PCA(S)$ and $X_{T_{root}} \gets PCA(T)$
\State $SubspaceAlign(X_{S_{root}},X_{T_{root}})$ or $GFK(X_{S_{root}},X_{T_{root}})$
\State $ClassifyParent$(Target Data T)
\State $i \gets 0$
\While{$i<$ $No.$ $of$ $Parents$}
\State $X_{S_{i}} \gets PCA(S_i)$ \Comment $S_i$s are labeled data points (from $i^{th}$ parent)
\State $X_{T_{i}} \gets PCA(T_i)$ \Comment $T_i$s are data points classified as $i^{th}$ parent by the root classifier
\State $SubspaceAlign(X_{S_{i}},X_{T_{i}})$ or $GFK(X_{S_{i}},X_{T_{i}})$
\State $ClassifyChild$(Target Data $T_i$)
\EndWhile\label{euclidendwhile}
\State \textbf{return} $Accuracy$
\EndProcedure
\end{algorithmic}
\end{algorithm}
\vspace{-0.3cm}
In hierarchical subspace alignment we learn different \textit{metric} $M$ at different levels of hierarchy independently. Without loss of generality, we consider here hierarchies with only two levels, i.e. composed of a root node, a set of parent nodes (each corresponding to a set of similar categories) and a set of child nodes or leaf nodes (corresponding to the different categories). Hence the mathematical formulation of our approach is governed by \textit{eq.} \ref{equ2} and \ref{equ3}. 
\vspace{-0.1cm}
\begin{eqnarray}
	F(M_{root}) = \|X_{S_{root}}M_{root}-X_{T_{root}}\|_F^2
	\qquad  M_{root}^* = \argmin_{M_{root}} (F(M_{root})) 
	\label{equ2}
\end{eqnarray}
\vspace{-0.5cm}
\begin{eqnarray}    
	\forall i \in parent, F(M_{i}) = \|X_{S_{i}}M_{i}-X_{T_{i}}\|_F^2
	\qquad  M_{i}^* = \argmin_{M_{i}} (F(M_{i})) 
	\label{equ3}
\end{eqnarray}
\vspace{-0.1cm}
Here $M_{root}^*$ is the transformation matrix learned at the topmost level of hierarchy to differentiate between the parents. Each parent category consists of several similar child categories. $X_{S_{root}}$ and $X_{T_{root}}$ are the source and target subspaces considering all the source and target data. $M_{i}^*$ is the transformation matrix learned at the second level of hierarchy to distinguish between the children of parent \textit{i}. $X_{S_{i}}$ and $X_{T_{i}}$ are source and target subspaces for categories that belong to parent \textit{i}. Hence $X_{S_{i}}$ and $X_{T_{i}}$ are obtained using only the data points that belong to a child category of parent \textit{i}. Solutions of the \textit{eq.} \ref{equ2} and \ref{equ3} are similar to \textit{eq.} \ref{equ1}.

In hierarchical geodesic flow kernel we compute different kernel matrices at different levels of hierarchy for categorization at a specific level. For classifying between the parent classes, kernel matrix $G_{root}$ is computed considering the source and target subspaces generated by all the data. For classifying between the children of a specific parent category $i$ we compute kernel matrix $G_i$ considering the subspace obtained from the children classes of parent $i$. Similarity between two data points depends on the hierarchy level at which prediction is performed. 

\vspace{-0.5cm}

\section{Experiments}
\label{experiments}
\vspace{-0.3cm}

\begin{figure}
        \centering
        \begin{subfigure}[b]{0.15\textwidth}
                \includegraphics[width=2cm , height =2cm]{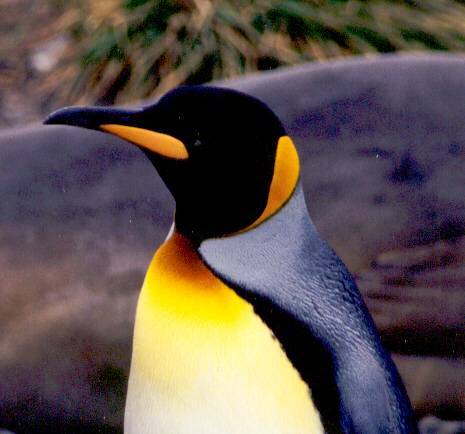}
                \caption{Penguin caltech 256}
        \end{subfigure}
        \begin{subfigure}[b]{0.15\textwidth}
                \includegraphics[width=2cm , height =2cm]{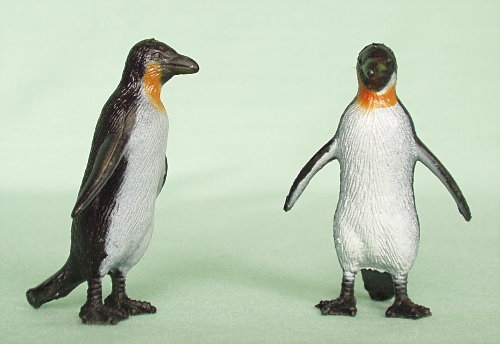}
                \caption{Penguin bing caltech}
        \end{subfigure}
        ~ 
        \begin{subfigure}[b]{0.15\textwidth}
                \includegraphics[width=2cm , height =2cm]{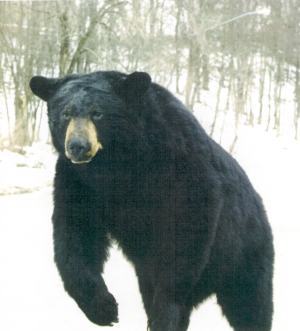}
                \caption{Bear caltech 256}
        \end{subfigure}
        \begin{subfigure}[b]{0.15\textwidth}
                \includegraphics[width=2cm , height =2cm]{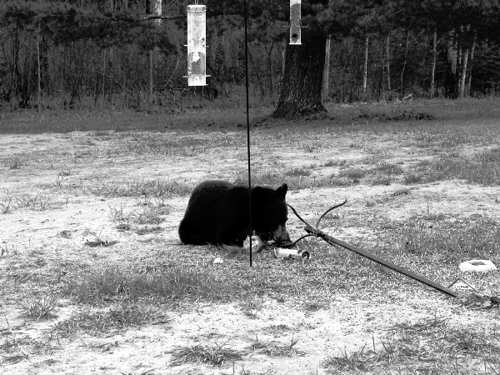}
                \caption{Bear bing caltech}
        \end{subfigure}
        ~ 
        \begin{subfigure}[b]{0.15\textwidth}
                \includegraphics[width=2cm , height =2cm]{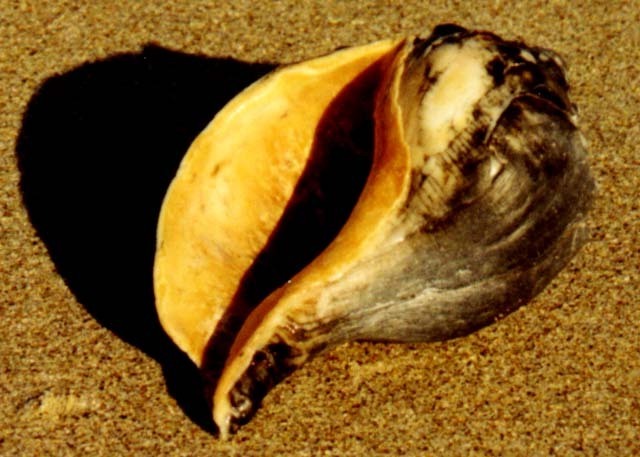}
                \caption{Conch caltech 256}
        \end{subfigure}
        \begin{subfigure}[b]{0.15\textwidth}
                \includegraphics[width=2cm , height =2cm]{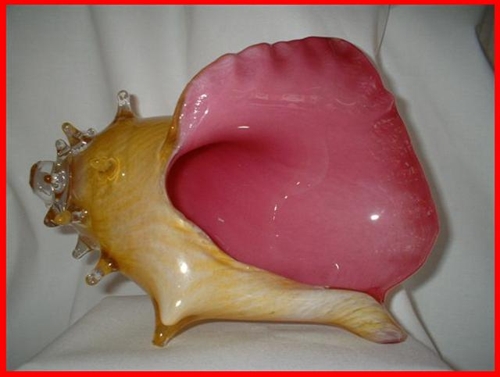}
                \caption{Conch bing caltech}
        \end{subfigure}
        \caption{Image samples taken from caltech-256 and bing caltech to show domain shift}\label{fig:animals}
\end{figure}

In this section we evaluate our results on a part of hierarchy taken from the caltech-256 and bing-caltech \cite{ES17,ES18}. We show our experimental results on the animal hierarchy consisting of the following three parent nodes: aquatic, terrestrial and avian animals and each parent consists of several child categories. For each image in the dataset we compute $4096-$ dimensional convolutional neural network based features obtained using Decaf \cite{ES16} by first resizing the full image to the desired input size. Note that the dataset has not been augmented with  any virtual examples by flipping or random cropping. In this paper we have used K-NN as our classifier as this has also been similarly used in \cite{ES1, ES2}. The rank of the domain is decided using the procedure given in \cite{ES2} and based on this procedure we fix the dimensionality of the subspaces for both root and parent subspaces as $53$. We first show the result without applying any domain adaptation algorithm  on the source and target data to show that there exists a non-negligible domain shift between these two datasets. This is shown in table \ref{tab2} in column ``Base Accuracy''. The results in table \ref{tab2} show that the hierarchy based subspace alignment consistently improves the results.  We also evaluate the similarity between subspaces of source and target domain by taking dot product ($trace(A'*B)$) at various levels of hierarchy to analyse our approach. This result is provided in table \ref{tab1}. As can be seen from the table \ref{tab1}, the maximum similarity is observed to be between the relevant subspaces in source and target. The low values we obtain off the main diagonal indicate the subspaces
for the different parent nodes are quite different from one another and different from the
root subspace.\\

\vspace{-0.3cm}
\begin{table}[ht!]

\centering
\vspace{-0.3cm}
    \begin{tabular}{|l|l|l|l|l|l|}
    \hline
        &  &  & Base & Accuracy & Accuracy    \\ [-1ex]
    \raisebox{1.5ex} {Source Dataset}&\raisebox{1.5ex} {Target Dataset} &\raisebox{1.5ex} {Method} & Accuracy & (without Hierarchy) & (with Hierarchy) \\ \hline
		
    Caltech-256         & Caltech-Bing    &GFK &    24.41      &     39.96        & \textbf{40.41}          \\
   
    Caltech-Bing  & Caltech-256  & GFK &      21.11       &       45.23        &  \textbf{49.67}     \\ 
        Caltech-256         & Caltech-Bing  & SA &   24.41         &   39.24         &  \textbf{40.78}           \\
   
    Caltech-Bing  & Caltech-256    & SA &   21.11       &   44.12        &  \textbf{48.36}     \\ \hline
    
    \end{tabular}

    \caption{Result for hierarchical domain adaptation on animal hierarchy of caltech-256 and bing-caltech. Here GFK represents Geodesic flow kernel and SA represents subspace alignment }
        \label{tab2}
\end{table}

\vspace{-0.3cm}

\begin{table}[ht!]

\centering
    \vspace{-0.3cm}
    \begin{tabular}{|l|l|l|l|l|}
    \hline
      & Root(Target)&Avian(Target)& Terrestrial(Target)&Aquatic(Target)  \\ \hline
      Root(Source) & \textbf{3.96} & 1.35 & 0.26 & 2.05 \\
      Avian(Source) & 3.11&\textbf{3.74} & -0.10  & 0.18 \\
      Terrestrial(Source) & 1.44 & 2.62 & \textbf{3.92} & 0.76 \\
      Aquatic(Source) & 1.22 & 0.43 & 1.42 & \textbf{3.16} \\ 
    \hline
    \end{tabular}

    \caption{Similarity Matrix for Source Subspace and Target Subspace considering each Hierarchy level separately. Here caltech-256 is considered as source and bing-caltech as target.}
        \label{tab1}
\end{table}

\vspace{-0.3cm}

\vspace{-0.5cm}
\section{Conclusion}
\label{Conclude}
\vspace{-0.4cm}

In this paper, we have considered a hierarchical subspace based domain adaptation approach. Based on the evaluation we observe that considering different domain adaptation subspaces specific to the individual category level can indeed aid the domain adaptation. In future, we would like to evaluate the effect of restricting the subspaces to groups of classes which need not be obtained strictly based on hierarchy which would generalize the approach to any source and target domains that are not hierarchically labeled.


\newpage

\begin{thebibliography}{9}

  \bibitem{ES1}
  Fernando, Basura, Amaury Habrard, Marc Sebban, and Tinne Tuytelaars. "Unsupervised visual domain adaptation using subspace alignment." In Computer Vision (ICCV), 2013 IEEE International Conference on, pp. 2960-2967. IEEE, 2013
  \bibitem{ES2}
  Gong, Boqing, Yuan Shi, Fei Sha, and Kristen Grauman. "Geodesic flow kernel for unsupervised domain adaptation." In Computer Vision and Pattern Recognition (CVPR), 2012 IEEE Conference on, pp. 2066-2073. IEEE, 2012.
  \bibitem{ES3}
  Gopalan, Raghuraman, Ruonan Li, and Rama Chellappa. "Domain adaptation for object recognition: An unsupervised approach." In Computer Vision (ICCV), 2011 IEEE International Conference on, pp. 999-1006. IEEE, 2011.
  
  \bibitem{ES4}
  Torralba, Antonio, and Alexei A. Efros. "Unbiased look at dataset bias." In Computer Vision and Pattern Recognition (CVPR), 2011 IEEE Conference on, pp. 1521-1528. IEEE, 2011.

  \bibitem{ES5}
  Margolis, Anna. "A literature review of domain adaptation with unlabeled data." Tec. Report (2011): 1-42.

  \bibitem{ES6}
  Uribe, Diego. "Domain adaptation in sentiment classification." In Machine Learning and Applications (ICMLA), 2010 Ninth International Conference on, pp. 857-860. IEEE, 2010.

  \bibitem{ES7}
  Glorot, Xavier, Antoine Bordes, and Yoshua Bengio. "Domain adaptation for large-scale sentiment classification: A deep learning approach." In Proceedings of the 28th International Conference on Machine Learning (ICML-11), pp. 513-520. 2011.

  \bibitem{ES8}
  Ben-David, Shai, John Blitzer, Koby Crammer, and Fernando Pereira. "Analysis of representations for domain adaptation." Advances in neural information processing systems 19 (2007): 137.

  \bibitem{ES9}
  Leggetter, Christopher J., and Philip C. Woodland. "Maximum likelihood linear regression for speaker adaptation of continuous density hidden Markov models." Computer Speech \& Language 9.2 (1995): 171-185.

  \bibitem{ES10}
  Nguyen, Hien V., Huy Tho Ho, Vishal M. Patel, and Rama Chellappa. "Joint hierarchical domain adaptation and feature learning." IEEE Transactions on Pattern Analysis and Machine Intelligence, submitted (2013).

  \bibitem{ES11}
  Xu, Jiaolong, Sebastian Ramos, David Vázquez, and Antonio M. López. "Hierarchical Adaptive Structural SVM for Domain Adaptation." arXiv preprint arXiv:1408.5400 (2014).

  \bibitem{ES12}
  Finkel, Jenny Rose, and Christopher D. Manning. "Hierarchical bayesian domain adaptation." Proceedings of Human Language Technologies: The 2009 Annual Conference of the North American Chapter of the Association for Computational Linguistics. Association for Computational Linguistics, 2009.
  
  \bibitem{ES13}
  Patel, Vishal M., Raghuraman Gopalan, Ruonan Li, and Rama Chellappa. "Visual Domain Adaptation: An Overview of Recent Advances."
  
  \bibitem{ES14}
  Khosla, Aditya, Tinghui Zhou, Tomasz Malisiewicz, Alexei A. Efros, and Antonio Torralba. "Undoing the damage of dataset bias." In Computer Vision–ECCV 2012, pp. 158-171. Springer Berlin Heidelberg, 2012.
  
  \bibitem{ES15}
  Perronnin, Florent, Jorge Sánchez, and Yan Liu. "Large-scale image categorization with explicit data embedding." In Computer Vision and Pattern Recognition (CVPR), 2010 IEEE Conference on, pp. 2297-2304. IEEE, 2010.
  \bibitem{ES16}
  Donahue, Jeff, Yangqing Jia, Oriol Vinyals, Judy Hoffman, Ning Zhang, Eric Tzeng, and Trevor Darrell. "DeCAF: A Deep Convolutional Activation Feature for Generic Visual Recognition." In Proceedings of The 31st International Conference on Machine Learning, pp. 647-655. 2014.
  \bibitem{ES17}
  Griffin, Gregory, Alex Holub, and Pietro Perona. "Caltech-256 object category dataset." (2007).
  \bibitem{ES18}
  Bergamo, Alessandro, and Lorenzo Torresani. "Exploiting weakly-labeled web images to improve object classification: a domain adaptation approach." In Advances in Neural Information Processing Systems, pp. 181-189. 2010.
\end{thebibliography}
\end{document}